\documentclass[runningheads]{llncs}

\makeatletter
\newcommand{\@chapapp}{\relax}%
\makeatother

\usepackage{graphicx}
\usepackage{amsmath}
\usepackage{hyperref}
\usepackage{dsfont}
\usepackage{amssymb}
\usepackage{upgreek}
\usepackage{verbatim}
\usepackage{subcaption}
\usepackage{algorithm}
\usepackage{algorithmic}

\usepackage[title,toc,titletoc,header]{appendix}
\DeclareMathOperator*{\argmax}{arg\,max}

\begin{document}
\title{Online Algorithms for Multiclass Classification using Partial Labels}
%
%
\author{Rajarshi Bhattacharjee \inst{1}
\and
Naresh Manwani\inst{2}}
\authorrunning{R. Bhattacharjee and N. Manwani}
%
\institute{IIT Madras, India\\
\email{brajarshi91@gmail.com}\\
 \and
IIIT Hyderabad, India \\
\email{naresh.manwani@iiit.ac.in}}
\maketitle              
\begin{abstract}
In this paper, we propose online algorithms for multiclass classification using partial labels. We propose two variants of Perceptron called Avg Perceptron and Max Perceptron to deal with the partial labeled data. We also propose Avg Pegasos and Max Pegasos, which are extensions of Pegasos algorithm. We also provide mistake bounds for Avg Perceptron and regret bound for Avg Pegasos. We show the effectiveness of the proposed approaches by experimenting on various datasets and comparing them with the standard Perceptron and Pegasos.

\end{abstract}
\section{Introduction}
Multiclass classification is a well-studied problem in machine learning. However, we assume that we know the true label for every example in the training data. In many applications, we don't have access to the true class label as labeling data is an expensive and time-consuming process. Instead, we get a set of candidate labels for every example. This setting is called multiclass learning with partial labels. The true or ground-truth label is assumed to be one of the instances in the partial label set. Partially labeled data is relatively easier to obtain and thus provides a cheap alternative to learning with exact labels.

Learning with partial labels is referred to as superset label learning \cite{Dietterich}, ambiguous label learning \cite{Chellappa}, and by other names in different papers. Many proposed models try to \textit{disambiguate} the correct labels from the incorrect ones. One popular approach is to treat the unknown correct label in the candidate set as a latent variable and then use an Expectation-Maximization type algorithm to estimate the correct label as well the model parameters iteratively (\cite{Ghahramani}, \cite{Vannoorenberghe}, \cite{Dietterich}, \cite{Bengio}, \cite{Chellappa}). Other approaches to label disambiguation include using a maximum margin formulation \cite{marginPL}, which alternates between ground truth identification and maximizing the margin from the ground-truth label to all other labels. Another model assumes that the ground truth label is the one to which the maximum score is assigned in the candidate label set by the model \cite{Caruana}. Then the margin between this ground-truth label and all other labels not in the candidate set is maximized. 

Some approaches try to predict the label of an unseen instance by averaging the candidate labeling information of its nearest neighbors in the training set (\cite{Yu}, \cite{Hullermeier}). Some formulations combine the partial label learning framework with other frameworks like multi-label learning \cite{Xie}. There are also specific approaches that do not try to disambiguate the label set directly. For example, Zhang et al. \cite{Zhang} introduced an algorithm that works to utilize the entire candidate label set using a method involving error-correcting codes.

A general risk minimization framework for learning with partial labels is discussed in Cour et al. (\cite{Taskar}, \cite{Cour}). In this framework, any standard convex loss function can be modified to be used in the partial label setting. For a single instance, since the ground-truth label is not available, an average over the scores in the candidate label set is taken as a proxy to calculate the loss. Nguyen and Caruana \cite{Caruana} propose a risk minimization approach based on a non-convex max-margin loss for a partial label setting. 

In this paper, we propose online algorithms for multiclass classification using partially labeled data. Perceptron \cite{Perceptron} algorithm is one of the earliest online learning algorithms. Perceptron for multiclass classification is proposed in \cite{Duda}. A unified framework for designing online update rules for multiclass classification was provided in \cite{Ultraconservative}. An online variant of the support vector machine \cite{Smola2004} called Pegasos is proposed in \cite{Pegasos}.  This algorithm is shown to achieve $O(\log T)$ \textit{regret} (where $T$ is the number of rounds). Once again, all these online approaches assume that we know the true label for each example.

Online multiclass learning with partial labels remained an unaddressed problem. In this paper, we propose several online multiclass algorithms using partial labels. Our key contributions in this paper are as follows.
\begin{enumerate}
\item We propose Avg Perceptron and Max Perceptron, which extensions of Perceptron to handle the partial labels. Similarly, we propose Avg Pagasos and Max Pegasos, which are extensions of Pegasos algorithm.
\item We derive mistake bounds for Avg Perceptron in both separable and general cases. Similarly, we provide $\log(T)$ regret bound for Avg Pegasos. 
\item We also provide thorough experimental validation of our algorithms using datasets of different dimensions and compare the performance of the proposed algorithms with standard multiclass Perceptron and Pegasos.
\end{enumerate}

\section{Multiclass Classification Using Partially Labeled Data}
We now formally discuss the problem of multiclass classification given partially labeled training set.
Let $ \mathcal{X} \subseteq \mathbb{R}^d $ be the feature space from which the instances are drawn and let $\mathcal{Y}=\{1,\ldots,K\}$ be the output label space. Every instance $\mathbf{x} \in \mathcal{X}$ is associated with a candidate label set $Y \subseteq \mathcal{Y}$. The set of labels not present in the candidate label set is denoted by $\overline{Y}$. Obviously, $Y \cup \overline{Y}=[K]$.\footnote{We denote the set $\{1,\ldots,K\}$ using $[K]$.} The ground-truth label associated with $\mathbf{x}$ is denoted by lowercase $y$. It is assumed that the actual label lies within the set $Y$ (i.e., $y \in Y$). The goal is to learn a classifier $h:\mathcal{X}\rightarrow \mathcal{Y}$. Let us assume that $h(\mathbf{x})$ is a linear classifier. Thus, $h(\mathbf{x})$ is parameterized by a matrix of weights $W \in \mathbb{R}^{d\times K}$ and is defined as $h(\mathbf{x})=\argmax_{i \in [K]}\;\; \mathbf{w}_i. \mathbf{x}$
where $\mathbf{w}_i$ ($i$th column vector of $W$) denotes the parameter vector corresponding to the $i^{th}$ class. Discrepancy between the true label and the predicted label is captured using 0-1 loss as $L_{0-1}(h(\mathbf{x}),y)=\mathbb{I}_{\{h(\mathbf{x}) \neq y\}}$.
Here, $\mathbb{I}$ is the 0-1 indicator function, which evaluates to true when the condition mentioned is true and 0 otherwise. However, in the case of partial labels, we use partial (ambiguous) 0-1 loss \cite{Taskar} as follows. 
\begin{equation}\label{ambiguous}
    L_{A}(h(\mathbf{x}),Y)=\mathbb{I}_{\{h(\mathbf{x}) \notin Y\}}
\end{equation}
Minimizing $L_{A}$ is difficult as it is not continuous. Thus, we use continuous surrogates for $L_A$. A convex surrogate of $L_A$ is the {\em average prediction} hinge loss (APH) \cite{Taskar} which is defined as follows.
\begin{equation}\label{Avgloss}
L_{APH}(h(\mathbf{x}),Y) = \left[1-\frac{1}{|Y|}\sum_{i \in Y}\mathbf{w}_i.\mathbf{x}+\max_{j \notin  Y} \mathbf{w}_j.\mathbf{x}\right]_+ 
\end{equation}
where $|Y|$ is the size of the candidate label set and $[a]_+=\max(a,0)$. $L_{APH}$ is shown to be a convex surrogate of $L_A$ in \cite{Cour}.
There is another non-convex surrogate loss function called the {\em max prediction} hinge loss (MPH) \cite{Caruana} that can be used for partial labels which is defined as follows:
\begin{equation}\label{Maxloss}
    L_{MPH}(h(\mathbf{x}),Y) = \left[1-\max_{i \in Y}\mathbf{w}_i.\mathbf{x}+\max_{j \notin  Y} \mathbf{w}_j.\mathbf{x}\right]_+
\end{equation}
In this paper, we present online algorithms based on based on stochastic gradient descent on $L_{APH}$ and $L_{MPH}$.

\section{Multiclass Perceptron using Partial Labels}
In this section, we propose two variants of multiclass Perceptron using partial labels. Let the instance observed at time $t$ be $\mathbf{x}^t$ and its corresponding label set be $Y^t$. The weight matrix at time $t$ is $W^t$ and the $i$th column of $W^t$ is denoted by $\mathbf{w}_i^t$. To update the weights, we propose two different schemes: (a) Avg Perceptron (using stochastic gradient descent on $L_{APH}$) and (b) Max Perceptron (using stochastic gradient descent on $L_{MPH}$). We use following sub-gradients of the $L_{APH}$ and $L_{MPH}$.
\begin{align}
\label{grad-aph}
    \nabla_{\mathbf{w}_k}L_{APH}&=\begin{cases}
    0, & \text{if } \frac{1}{|Y|}\sum_{i \in Y}\mathbf{w}_i.\mathbf{x}-\max_{j \in  \overline{Y}}\mathbf{w}_j.\mathbf{x} \geq 1 \\
    -\frac{\mathbf{x}}{|Y|}, & \text{if } \frac{1}{|Y|}\sum_{i \in Y}\mathbf{w}_i.\mathbf{x}-\max_{j \in  \overline{Y}}\mathbf{w}_j.\mathbf{x} < 1 \\
     & \text{ and } k \in Y \\
    \mathbf{x}, & \text{if } \frac{1}{|Y|}\sum_{i \in Y}\mathbf{w}_i.\mathbf{x}-\max_{j \in  \overline{Y}}\mathbf{w}_j.\mathbf{x} < 1 \\ &   \text{ and } k=\argmax_{j \in \overline{Y}}\mathbf{w}_j.\mathbf{x} \\
    0, & \text{if } \frac{1}{|Y|}\sum_{i \in Y}\mathbf{w}_i.\mathbf{x}-\max_{j \in  \overline{Y}}\mathbf{w}_j.\mathbf{x} < 1 \\
    & \text{, } k \in \overline{Y}  \text{ and } k \neq \argmax_{j \in \overline{Y}}\mathbf{w}_j.\mathbf{x}
    \end{cases}
    \end{align}
    \begin{align}
    \label{grad-mph}
    \nabla_{\mathbf{w}_k}L_{MPH}&=\begin{cases}
    0, & \text{if } \max_{j \in  Y}\mathbf{w}_j.\mathbf{x}-\max_{j \in  \overline{Y}}\mathbf{w}_j.\mathbf{x} \geq 1 \\
    -\mathbf{x}, & \text{if } \max_{j \in  Y}\mathbf{w}_j.\mathbf{x}-\max_{j \in  \overline{Y}}\mathbf{w}_j.\mathbf{x} < 1 \\
     & \text{ and } k =\argmax_{i \in Y} \mathbf{w}_i.\mathbf{x} \\
    \mathbf{x}, & \text{if } \max_{j \in  Y}\mathbf{w}_j.\mathbf{x}-\max_{j \in  \overline{Y}}\mathbf{w}_j.\mathbf{x} < 1 \\ & \text{and } k = \argmax_{i \in \overline{Y}} \mathbf{w}_i.\mathbf{x} 
    \end{cases}
\end{align}
We initialize the weight matrix as a matrix of zeros. At trial $t$, the update rule for $\mathbf{w}_i$ can be written as:
\begin{align*}
    \mathbf{w}_i^{t+1}=\mathbf{w}_i^t-\eta\nabla_{\mathbf{w}_i}L(h^t(\mathbf{x}^t),Y^t)
\end{align*}
where $\eta>0$ is the step size and $\nabla_{\mathbf{w}_i}L(h^t(\mathbf{x}^t),Y^t)$ is found using Eq.(\ref{grad-aph}) and (\ref{grad-mph}). The complete description of Avg Perceptron and Max Perceptron is provided in Algorithm~\ref{Algo:AvgP} and \ref{Algo:MaxP} respectively. 

\begin{algorithm}
\caption{Avg Perceptron}
\label{Algo:AvgP}
\begin{algorithmic}
\STATE Initialize $W^1=0$ 
\FOR{$t=1$ to T} 
    \STATE Get $\mathbf{x}^t$
    \STATE Predict $\hat{y}^t$ as $\hat{y}^t=\argmax_{i \in [K]} \mathbf{w}_i^t.\mathbf{x}^t$
    \STATE Get the partial label set $Y^t$ of $\mathbf{x}^t$\\
    \STATE Calculate loss $L_{APH}(h^t(\mathbf{x}^t),Y^t)$ using Eq.(\ref{Avgloss})
    \IF{$L_{APH}(h^t(\mathbf{x}^t,Y^t)>0$}
        \STATE $\mathbf{w}_i^{t+1}=\mathbf{w}_i^t+\eta \tau_i^t\mathbf{x}^t,\;i \in [K]$ where
        \begin{align*}
            \tau_i^t=\begin{cases}
            \frac{1}{|Y^t|}, & i \in Y^t\\
            -1, & i=\argmax_{j \in \overline{Y}^t}\mathbf{w}_j^t.\mathbf{x}^t\\
            0, & \forall i\in \overline{Y}^t,\;i\neq \argmax_{j \in \overline{Y}^t}
            \end{cases}
        \end{align*}
    \ELSE
        \STATE $\mathbf{w}_i^{t+1} = \mathbf{w}_i^t,\;\forall i\in [K]$
    \ENDIF
\ENDFOR
\end{algorithmic}
\end{algorithm}

\begin{algorithm}[h]
\caption{Max Perceptron}
\label{Algo:MaxP}
\begin{algorithmic}
\STATE Initialize $W^1=0$ 
\FOR{$t=1$ to T} 
    \STATE Get $\mathbf{x}^t$
    \STATE Predict $\hat{y}^t$ as $\hat{y}^t=\argmax_{i \in [K]} \mathbf{w}_i^t.\mathbf{x}^t$
    \STATE Get the partial label set $Y^t$ of $\mathbf{x}^t$\\
    \STATE Calculate loss $L_{MPH}(h^t(\mathbf{x}^t),Y^t)$ using Eq.(\ref{Maxloss})
    \IF{$L_{MPH}(h^t(\mathbf{x}^t,Y^t)>0$}
        \STATE $\mathbf{w}_i^{t+1}=\mathbf{w}_i^t+\eta \tau_i^t\mathbf{x}^t,\;i \in [K]$ where
        \begin{align*}
            \tau_i^t=\begin{cases}
    1, & \text{if } \max_{j \in  Y}\mathbf{w}_j.\mathbf{x}-\max_{j \in  \overline{Y}}\mathbf{w}_j.\mathbf{x} < 1 \\
     & \text{ and } i =\argmax_{j \in Y} \mathbf{w}_j.\mathbf{x} \\
    -1, & \text{if } \max_{j \in  Y}\mathbf{w}_j.\mathbf{x}-\max_{j \in  \overline{Y}}\mathbf{w}_j.\mathbf{x} < 1 \\ & \text{and } i = \argmax_{j \in \overline{Y}} \mathbf{w}_j.\mathbf{x} 
    \end{cases}
        \end{align*}
    \ELSE
        \STATE $\mathbf{w}_i^{t+1} = \mathbf{w}_i^t,\;\forall i\in [K]$
    \ENDIF
\ENDFOR
\end{algorithmic}
\end{algorithm}
%

\subsection{Mistake Bound Analysis}
In the partial label setting, we say that mistake happens when the predicted class label for an example does not belong to its partial label set. We first define two variants of linear separability in a partial label setting as follows.

\begin{definition}[Average Linear Separability in Partial Label Setting]
Let $\{(\mathbf{x}^1,Y^1),$ \ldots, $(\mathbf{x}^T,Y^T)\}$ be the training set for multiclass classification with partial labels. We say that the data is {\it average linearly separable} if there exist $\mathbf{w}_1,\ldots,\mathbf{w}_K \in \mathbb{R}^{d}$ such that 
$$\frac{1}{|Y^t|}\sum_{i \in Y^t}\mathbf{w}_i.\mathbf{x}^t-\max_{j \in  \overline{Y}^t}\mathbf{w}_j.\mathbf{x}^t \geq \gamma,\;\forall t\in[T].$$
\end{definition}
Thus, average linear separability implies that $L_{APH}(h(\mathbf{x}^t),Y^t)=0,\;\forall t\in [T]$.

\begin{definition}[Max Linear Separability in Partial Label Setting]
Let $\{(\mathbf{x}^1,Y^1),$ \ldots, $(\mathbf{x}^T,Y^T)\}$ be the training set for multiclass classification with partial labels. We say that the data is {\it max linearly separable} if there exist $\mathbf{w}_1,\ldots,\mathbf{w}_K \in \mathbb{R}^{d}$ such that 
$$\max_{i \in Y^t}\mathbf{w}_i.\mathbf{x}^t-\max_{j \in  \overline{Y}^t}\mathbf{w}_j.\mathbf{x}^t \geq \gamma,\;\forall t\in[T].$$
\end{definition}
Thus, max linear separability implies that $L_{MPH}(h(\mathbf{x}^t),Y^t)=0,\;\forall t\in [T]$.

We bound the number of mistakes made by Avg Perceptron (Algorithm~\ref{Algo:AvgP}) as follows. 
\begin{theorem}[Mistake Bound for Avg Perceptron Under Average Linear Separability]\label{linear_mistake}
Let $(\mathbf{x}^1,Y^1),\ldots,(\mathbf{x}^T,Y^T)$ be the examples presented to Avg Perceptron, where $\mathbf{x}^t \in \mathbb{R}^d$ and $Y^t \subseteq [K]$. Let $W^* \in \mathbb{R}^{d \times K}$ ($\Vert W^* \Vert=1)$ be such that $ \frac{1}{|Y^t|}\sum_{i \in Y^t}\mathbf{w}^{*}_i.\mathbf{x}^t-\max_{j \in  \overline{Y}^t}\mathbf{w}^{*}_j.\mathbf{x}^t \geq \gamma,\;\forall t\in[T]$. Then we get the following mistake bound for Avg Perceptron Algorithm.
\begin{equation}
    \sum_{t=1}^T L_A(h^t(\mathbf{x}^t),Y^t) \leq\frac{2}{\gamma^2}+\left[\frac{1}{c}+1\right]\frac{R^2}{\gamma^2} \nonumber
\end{equation}
where $c=\min_t |Y^t|$, $R=\max_t ||\mathbf{x}^t||$ and $\gamma \geq 0$ is the margin of separation.
\end{theorem}
The proof is given in Appendix~\ref{proof-thm1}. We first notice that the bound is inversely proportional to the minimum label set size. This is intuitively obvious as the smaller the candidate label set size, the larger the chance of having a non-zero loss. When $c=1$, the number of updates reduces to the normal multiclass Perceptron mistake bound for linearly separable data as given in \cite{Ultraconservative}. Also, the number of mistakes is inversely proportional to $\gamma^2$. Linear separability (Definition~1) may not always hold for the training data. Thus, it is important to see how does the algorithm Avg Perceptron performs in such cases.
We now bound the number of updates in $T$ rounds for partially labeled data, which is linearly non-separable under $L_{APH}$. 

\begin{theorem}[Mistake Bound for Avg Perceptron in Non-Separable Case]\label{non-seprable}
Let $(\mathbf{x}^1,Y^1),\ldots,(\mathbf{x}^T,Y^T)$ be an input sequence presented to Avg Perceptron. Let $W$ ($\Vert W \Vert=1$) be weight matrix corresponding to a multiclass classifier. Then for a fixed $\gamma>0$, let $d^t=\max\left\{0,\gamma-[\frac{1}{|Y^t|}\sum_{i \in Y^t}\mathbf{w}_i.\mathbf{x}^t-\max_{j \in  \overline{Y}^t}\mathbf{w}_j.\mathbf{x}^t]\right\}$.
Let $D^2=\sum_{t=1}^T(|Y^t|d^t)^2$ and $R=\max_{t\in[T]} ||\mathbf{x}^t||$ and $c=\min_{t\in [T]} |Y^t|$. Then, mistakes bound for Avg Perceptron is as follows.
\begin{equation}
     \sum_{t=1}^T L_A(h^t(\mathbf{x}^t),Y^t)\leq 2\frac{Z^2}{\gamma^2}+2K\frac{R^2+\Updelta^2}{(\frac{\gamma}{Z})^2} \nonumber
\end{equation}
where $Z=\sqrt{1+\frac{D^2}{\Updelta^2}}$, $\Updelta=\left[\frac{D^2+KD^2R^2}{K}\right]^{\frac{1}{4}}$ and $K=\left[\frac{1}{c}+1\right]$.
\end{theorem}
The proof is provided in the Appendix~\ref{proof-thm2}.

\section{Online Multiclass Pegasos Using Partial Labels}
Pegasos \cite{Pegasos} is an online algorithm originally proposed for an exact label setting. In Pegasos, $L_2$ regularizer of the weights is minimized along with the hinge loss, making the overall objective function strongly convex. The strong convexity enables the algorithm to achieve a $O(\log T)$ regret in $T$ trials. The objective function of the Pegasos at trial $t$ is the following.
\begin{equation}
    f(W,\mathbf{x}^t,Y^t)=\frac{\lambda}{2}||W||^2 + L(h(\mathbf{x}^t),Y^t) \nonumber
\end{equation}
Here, $\lambda$ is a regularization constant and $||W||$ is Frobenius norm of the weight matrix. Let $W^t$ be the weight matrix at the beginning of trial $t$. Then, $W^{t+1}$ is found as $W^{t+1}=\Uppi_B(W^t-\eta_t\nabla^t)$.
Here $\nabla^t=\nabla_{W^t}f(W^t,\mathbf{x}^t,Y^t)$, $\eta_t$ is the step size at trial $t$ and $\Uppi_B$ is a projection operation onto the set $B$ which is defined as $B=\{W: ||W||\leq \frac{1}{\sqrt{\lambda}}\}$.
Thus, $\Uppi_B(W)=\min\{1,\frac{1}{(\lambda||W||)}\}W$.

We now propose extension of Pegasos \cite{Pegasos} for online multiclass learning using partially labeled data. We again propose two variants of Pegasos: (a) Avg Pegasos (using average prediction hinge loss (Eq.\ref{Avgloss})) and (b) Max Pegasos (using max prediction hinge loss (Eq.(\ref{Maxloss})). We first note that $\nabla^t$ can be written as:
\begin{equation}\label{pegasos_grad}
    \nabla^t=\lambda W^t+\nabla_{W^t}L
\end{equation}
where $\nabla_{W^t}L$ is given by Eq.(\ref{grad-aph}) (for $L_{APH}$) and Eq.(\ref{grad-mph}) (for $L_{MPH}$). Complete description of Avg Pegasos and Max Pegasos are given in Algorithm~\ref{algo:AvgPeg} and Algorithm~\ref{algo:MaxPeg} respectively.

\begin{algorithm}
\caption{Avg Pegasos}
\label{algo:AvgPeg}
\begin{algorithmic}
\STATE {\bf Input: }$\lambda,T$
\STATE {\bf Initialize: }$W_1$ s.t. $||W^1|| \leq \frac{1}{\sqrt{\lambda}}$
\FOR{$t = 1$ to $T$}
    \STATE Get $\mathbf{x}^t,Y^t$
    \STATE Set $\eta_t=\frac{1}{\lambda t}$
    \STATE Calculate loss 
    $L_{APH}(h^t(\mathbf{x}^t),Y^t)$ using Eq.(\ref{Avgloss})
    \IF{$L_{APH}>0$}
        \STATE $W^{t+\frac{1}{2}}=(1-\eta_t\lambda)W^t-\eta_t\nabla_{W}L_{APH}$ where $\nabla_{W}L_{APH}$ is given 
 by Eq.(\ref{grad-aph})
        \STATE $W^{t+1}=\min\{1,\frac{1/\sqrt{\lambda}}{||W^{t+\frac{1}{2}}||}\}W^{t+\frac{1}{2}}$
    \ELSE
        \STATE $W^{t+1}=W^t$
    \ENDIF
\ENDFOR
\STATE {\bf Output:} $W^T$
\end{algorithmic}
\end{algorithm}

\begin{algorithm}
\caption{Max Pegasos}
\label{algo:MaxPeg}
\begin{algorithmic}
\STATE {\bf Input: }$\lambda,T$
\STATE {\bf Initialize: }$W_1$ s.t. $||W^1|| \leq \frac{1}{\sqrt{\lambda}}$
\FOR{$t = 1$ to $T$}
    \STATE Get $\mathbf{x}^t,Y^t$
    \STATE Set $\eta_t=\frac{1}{\lambda t}$
    \STATE Calculate loss 
    $L_{MPH}(h^t(\mathbf{x}^t),Y^t)$ using Eq.(\ref{Maxloss})
    \IF{$L_{APH}>0$}
        \STATE $W^{t+\frac{1}{2}}=(1-\eta_t\lambda)W^t-\eta_t\nabla_{W}L_{MPH}$ where $\nabla_{W}L_{MPH}$ is given 
 by Eq.(\ref{grad-mph})
        \STATE $W^{t+1}=\min\{1,\frac{1/\sqrt{\lambda}}{||W^{t+\frac{1}{2}}||}\}W^{t+\frac{1}{2}}$
    \ELSE
        \STATE $W^{t+1}=W^t$
    \ENDIF
\ENDFOR
\STATE {\bf Output:} $W^T$
\end{algorithmic}
\end{algorithm}

\subsection{Regret Bound Analysis of Avg Pegasos}
We now derive the regret bound for Avg Pegasos. 
\begin{theorem}
Let $(\mathbf{x}^1,Y^1),(\mathbf{x}^2,Y^1),\ldots,(\mathbf{x}^T,Y^T)$ be an input sequence where $\mathbf{x}^t \in \mathbb{R}^d$ and $Y^t \subseteq [K]$. Let $R=\max_t ||\mathbf{x}^t||$.
Then the regret of Avg Pegasos is given as:
\begin{equation}
    \frac{1}{T}\sum\limits_{t=1}^T f(W^t,\mathbf{x}^t,Y^t) - \min_{W}\frac{1}{T}\sum\limits_{t=1}^T f(W,\mathbf{x}^t,Y^t) \leq \frac{G^2lnT}{\lambda T} \nonumber
\end{equation}
where $G=\sqrt{\lambda}+\sqrt{1+\frac{1}{c}}R$ and $c=\min_t |Y^t|$
\end{theorem}
The proof is given in Appendix~\ref{proof-thm3}. We again see the regret is inversely proportional to the size of the minimum candidate label set.

\section{Experiments}
We now describe the experimental results. 
We perform experiments on Ecoli, Satimage, Dermatology, and USPS datasets (available on UCI repository \cite{Dua:2019}) and MNIST dataset \cite{mnist}.
We perform experiments using the proposed algorithms Avg Perceptron, Max Perceptron, Avg Pegasos, and Max Pegasos. 
For benchmarking, we use Perceptron and Pegasos based on exact labels.

\begin{figure}[t]
    \centering
    \begin{subfigure}[t]{0.48\linewidth}
    \centering
    \includegraphics[width=\linewidth]{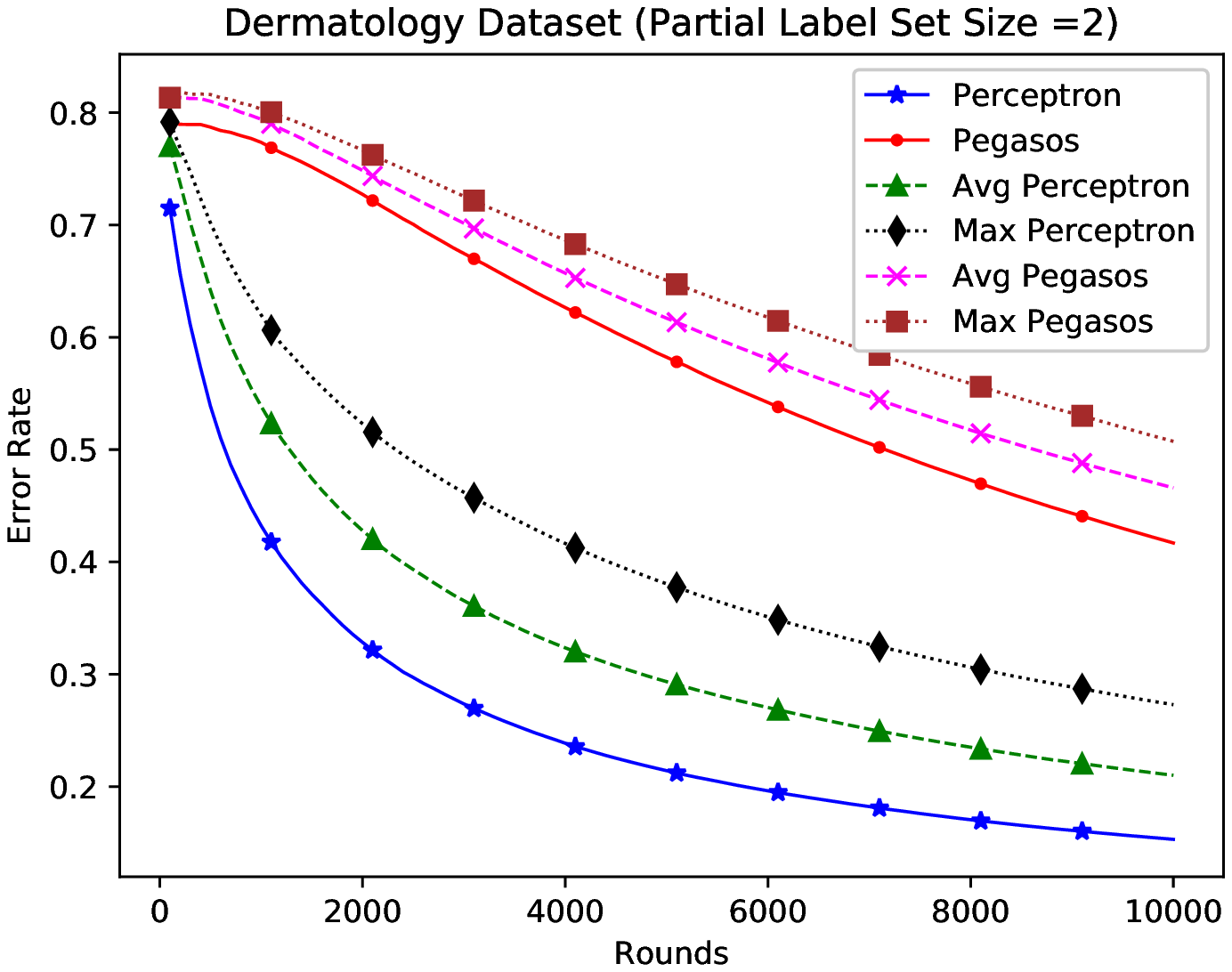}
    \end{subfigure}
    \begin{subfigure}[t]{0.48\linewidth}
    \centering
    \includegraphics[width=\linewidth]{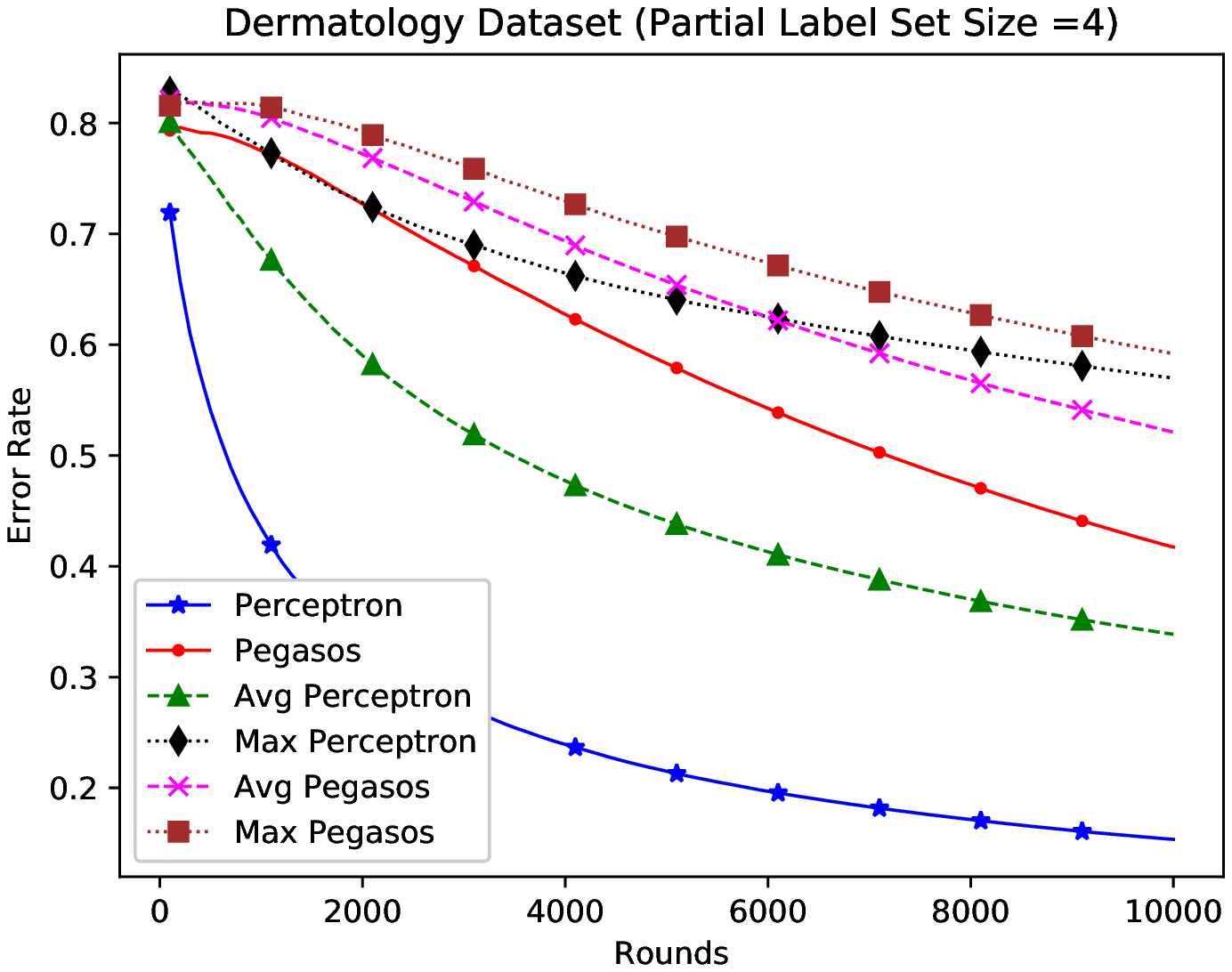}
    \end{subfigure}
    \caption{Dermatology Dataset Results}
    \label{fig:dermatology}
\end{figure}
\begin{figure}[t]
\begin{center}
\begin{tabular}{cc}
    \includegraphics[scale=0.4]{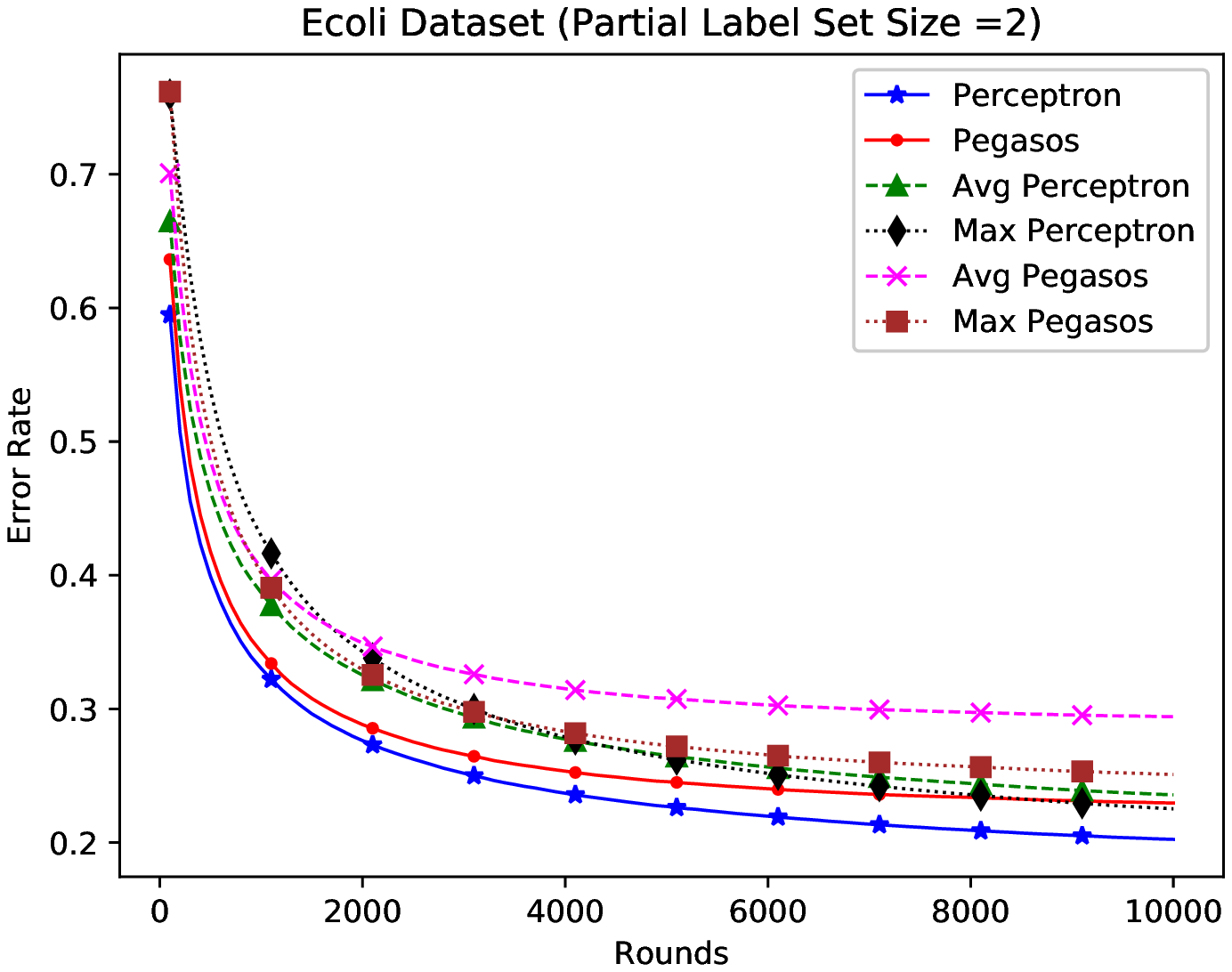} &
    \includegraphics[scale=0.4]{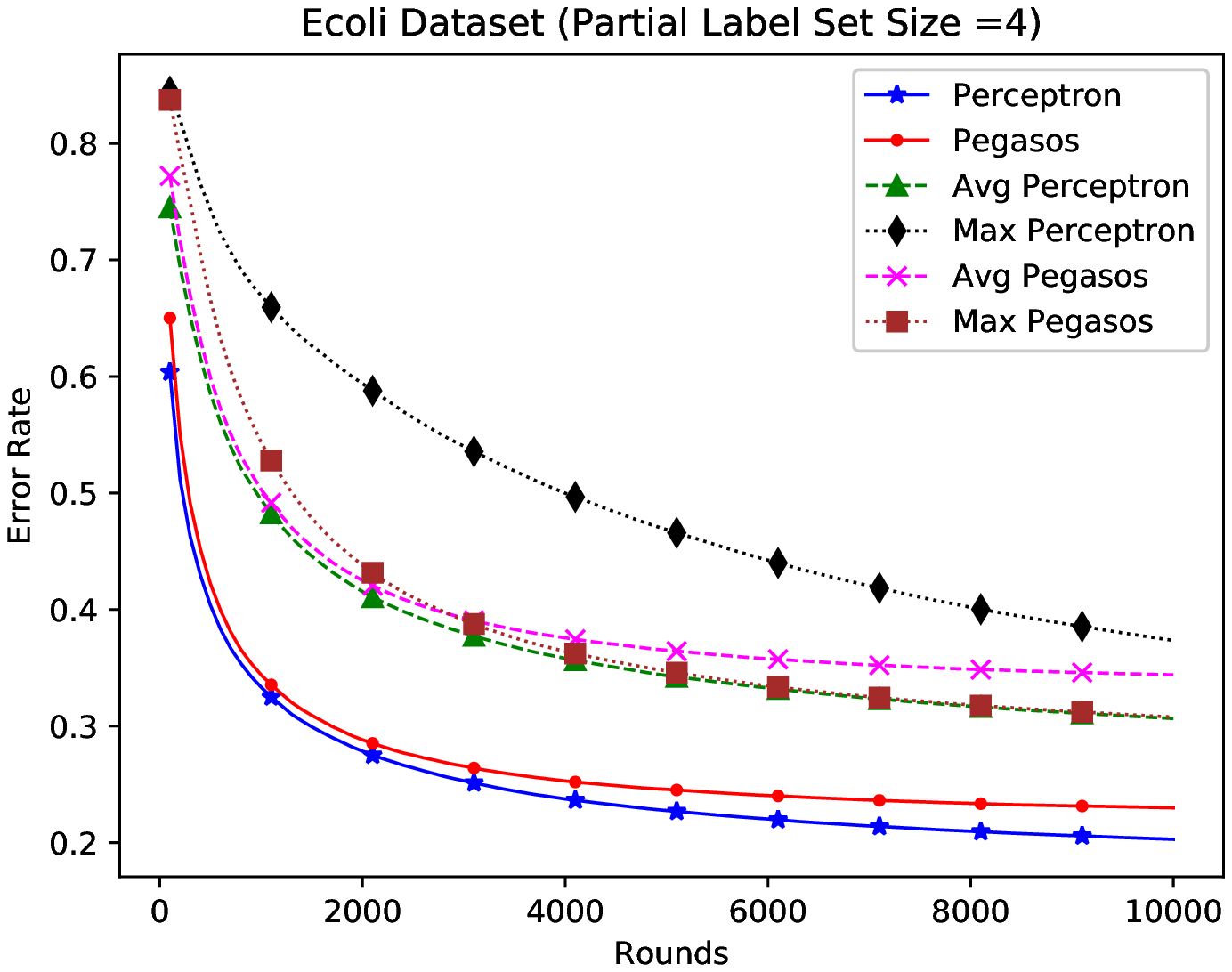}\\
    \includegraphics[scale=0.4]{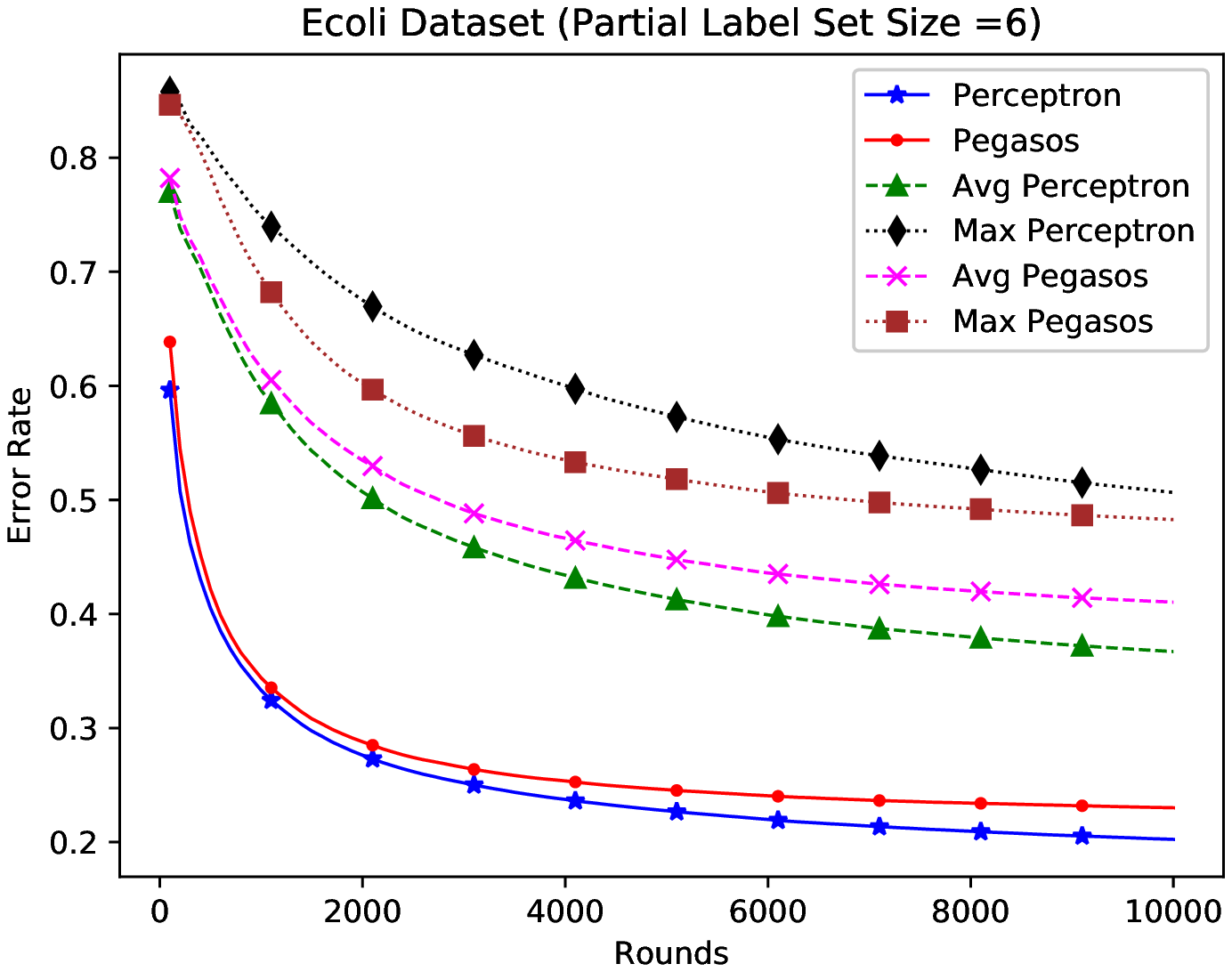} &
    \end{tabular}
    \caption{Ecoli Dataset Results}
    \label{fig:ecoli}
    \end{center}
\end{figure}
\begin{figure}[t]
    \centering
    \begin{subfigure}[t]{0.48\linewidth}
    \centering
    \includegraphics[width=\linewidth]{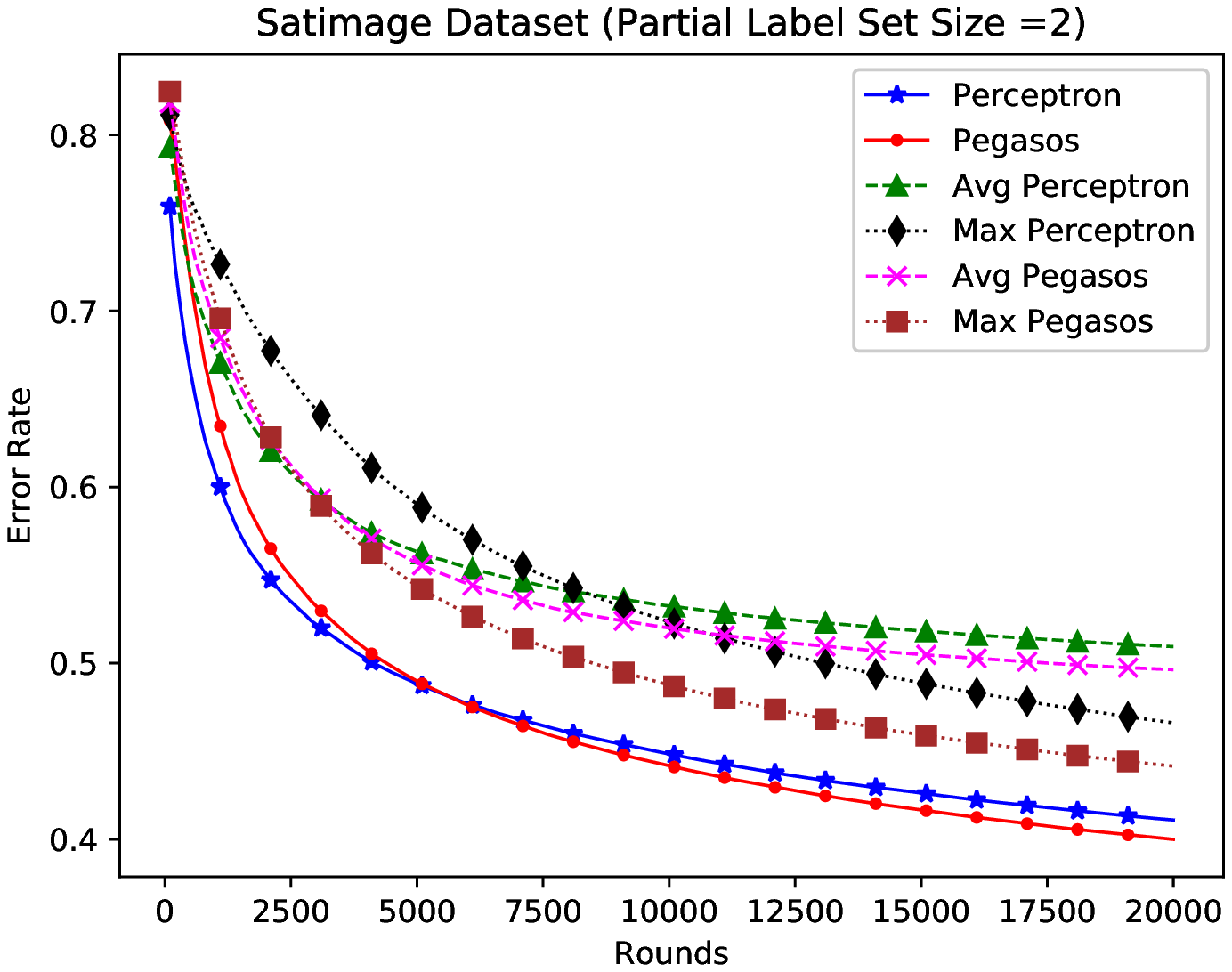}
    \end{subfigure}
    \begin{subfigure}[t]{0.48\linewidth}
    \centering
    \includegraphics[width=\linewidth]{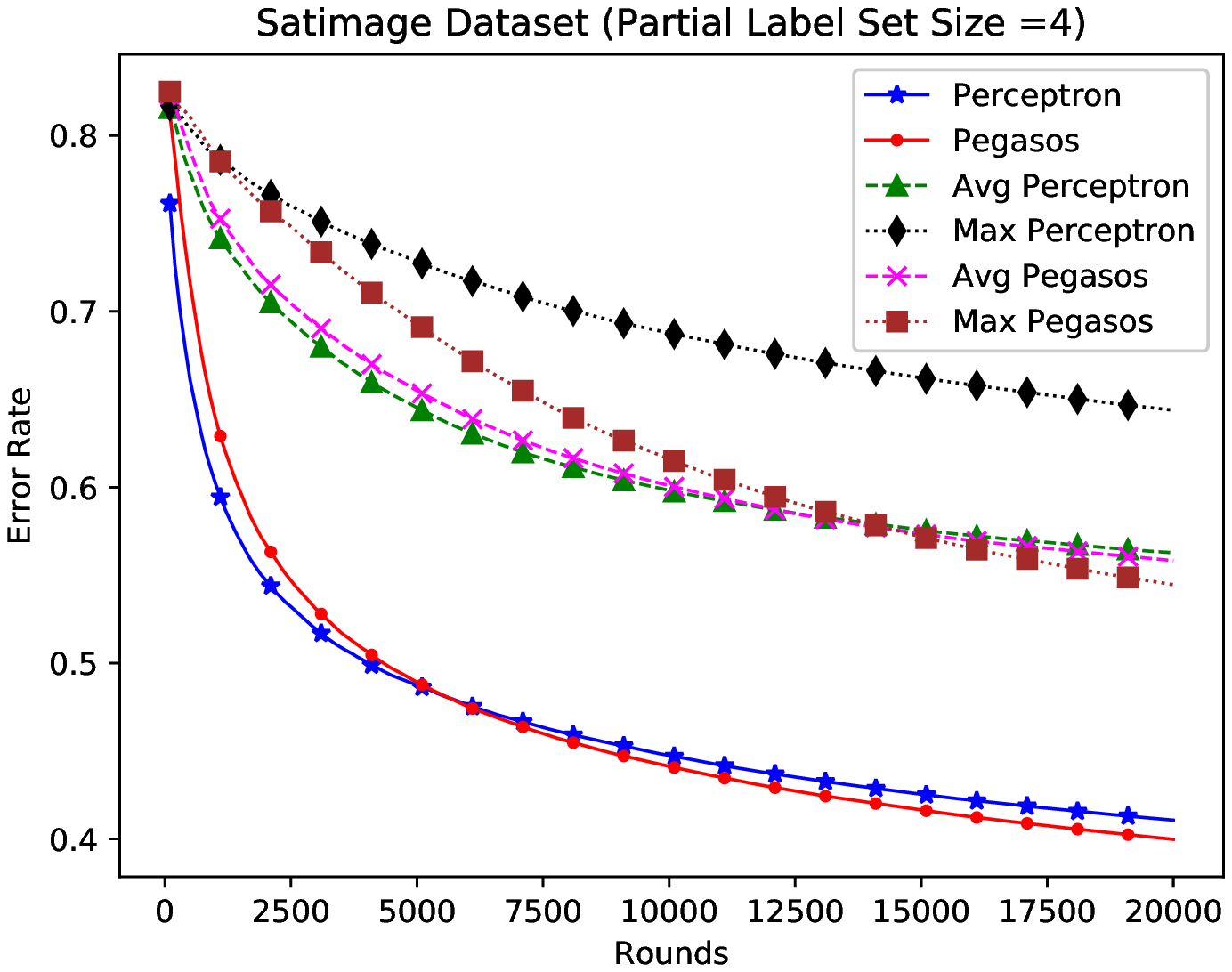}
    \end{subfigure}
    \caption{Satimage Dataset Results}
    \label{fig:satimage}
\end{figure}
\begin{figure*}[!t]
    \centering
    \begin{subfigure}[t]{0.48\linewidth}
    \centering
    \includegraphics[width=\linewidth]{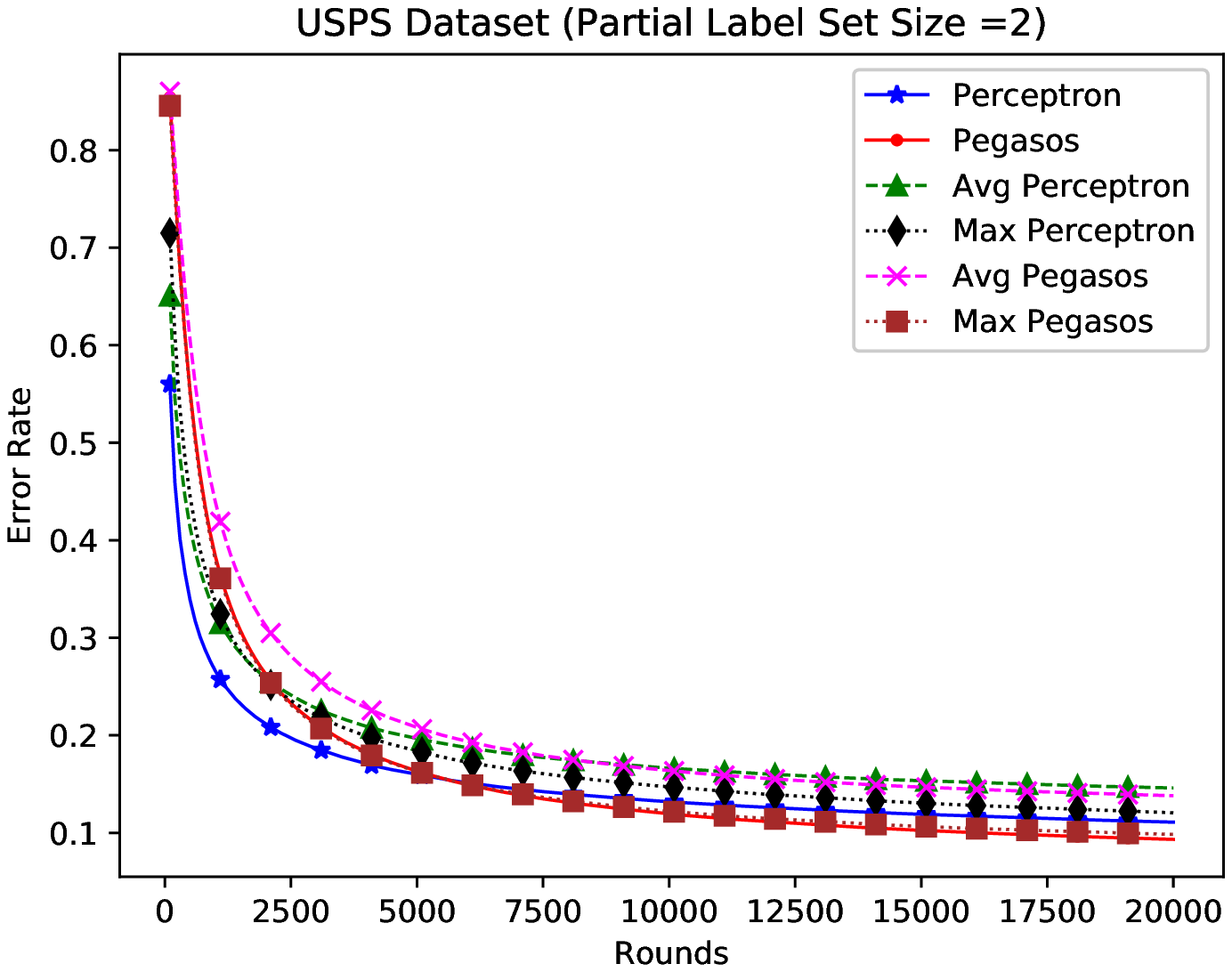}
    \end{subfigure}
    \begin{subfigure}[t]{0.48\linewidth}
    \centering
    \includegraphics[width=\linewidth]{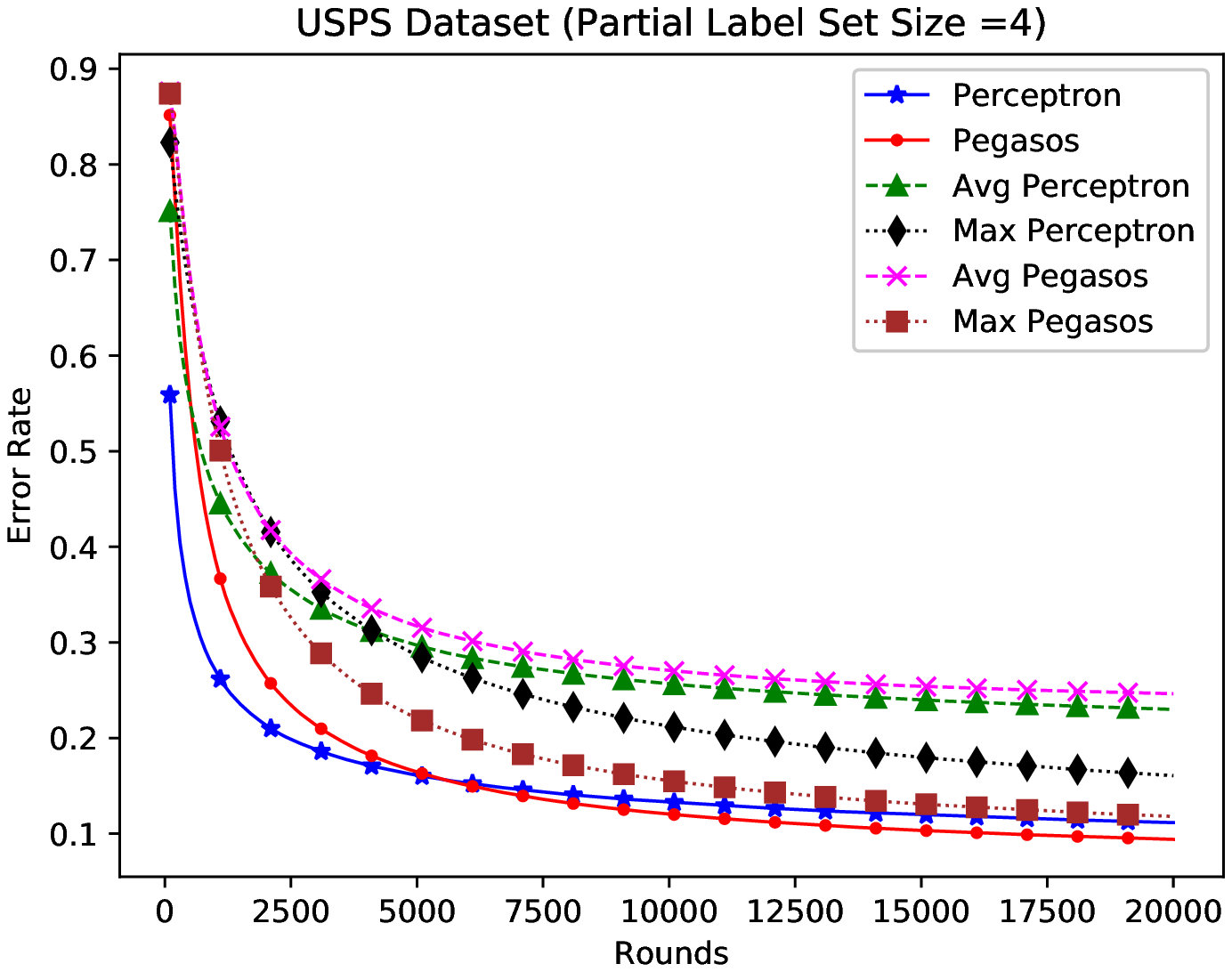}
    \end{subfigure}
    \begin{subfigure}[t]{0.48\linewidth}
    \centering
    \includegraphics[width=\linewidth]{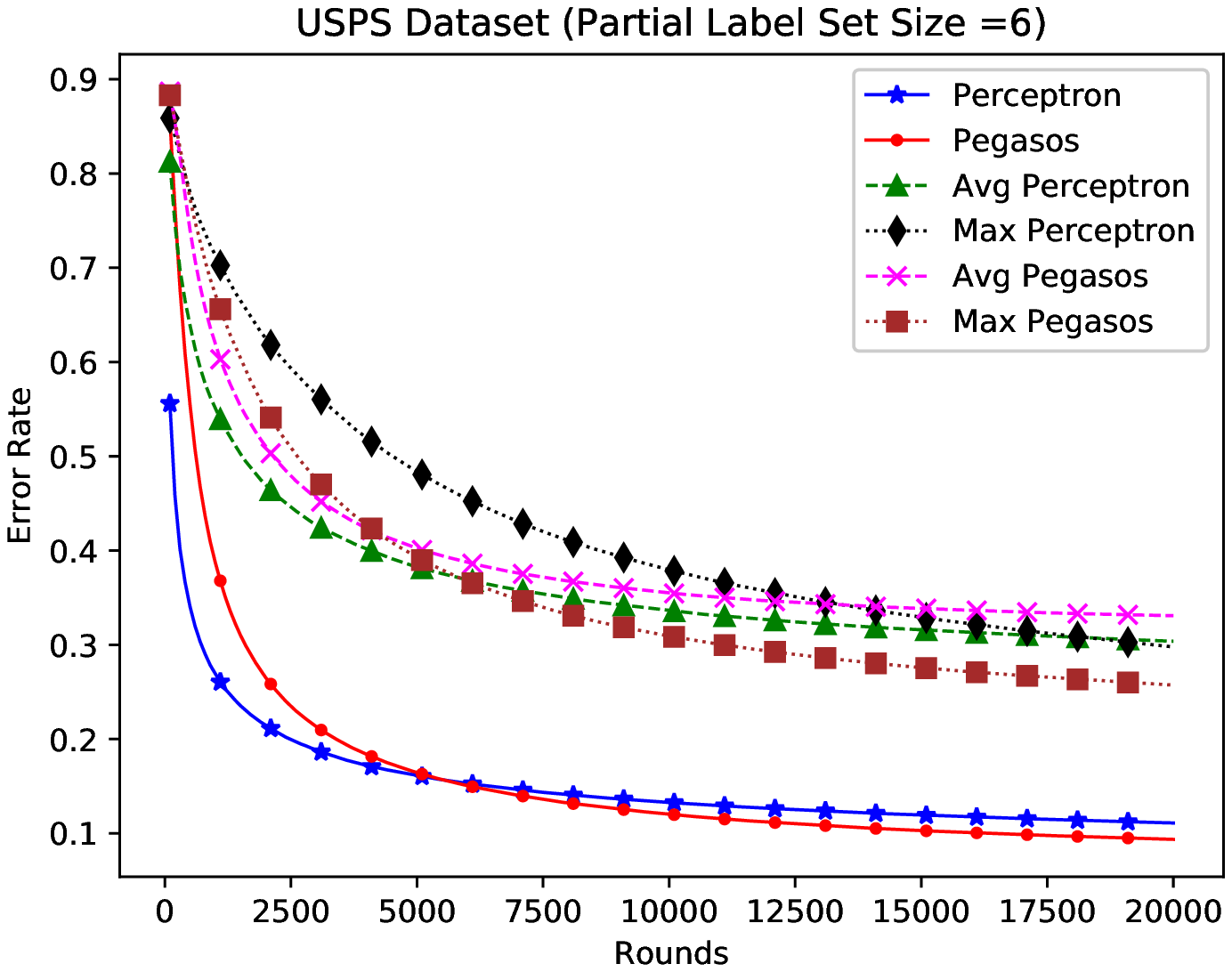}
    \end{subfigure}
    \begin{subfigure}[t]{0.48\linewidth}
    \centering
    \includegraphics[width=\linewidth]{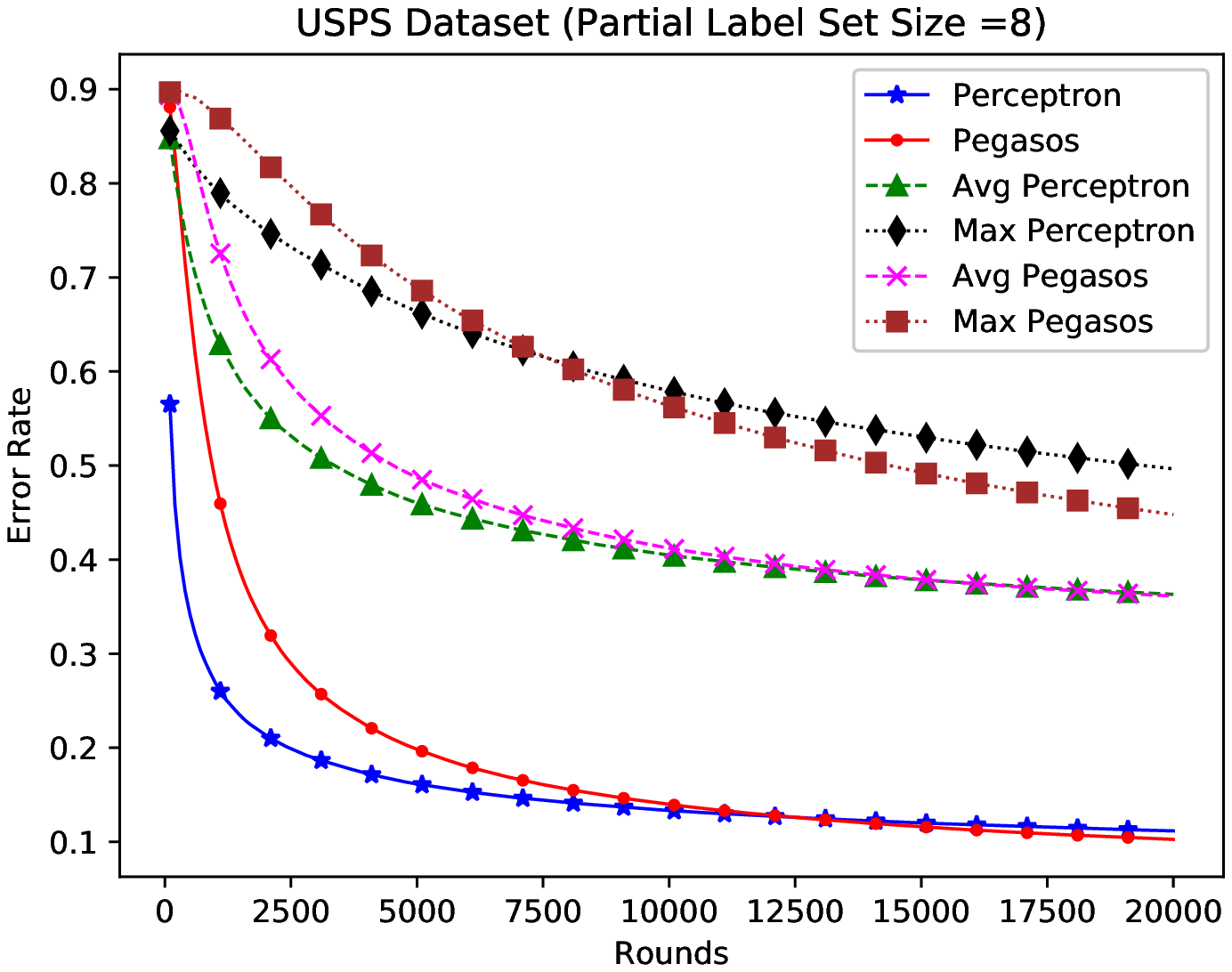}
    \end{subfigure}
    \caption{USPS Dataset Results}
    \label{fig:usps}
\end{figure*}
For all the datasets, the candidate or partial label set for each instance contains the true label and some labels selected uniformly at random from the remaining labels. After every trial, we find the average mis-classification rate (average of $L_{0-1}$ loss over examples seen till that trial) is calculated with respect to the true label. This sets a hard evaluation criteria for the algorithms. The number of rounds for each dataset is selected by observing when the error curves start to converge. For every dataset, we repeat the process of generating partial label sets and plotting the error curves 100 times and average the instantaneous error rates across the 100 runs. The final plots for each dataset have the average instantaneous error rate on the Y-axis and the number of rounds on the X-axis.

For every dataset, we plot the error rate curves for all the algorithms for different candidate label set sizes. This helps us in understanding how the online algorithms behave as the candidate label set size increases. For the Dermatology dataset, which contains six classes, we take candidate labels sets of sizes 2 and 4, respectively, as shown in Fig. \ref{fig:dermatology}. We see that the average prediction loss based algorithms perform the better in both cases. The results for the Ecoli dataset for candidate label sets of size 2,4 and 6 are shown in Fig. \ref{fig:ecoli}. Here, we find that the Max Pegasos algorithm performs comparably to the algorithms based on the Average Prediction Loss for candidate labels set sizes 2 and 4. But for candidate label set size 8, the Max Prediction Loss performs significantly worse than the Average Prediction Loss based algorithm.
\begin{figure*}[!t]
    \centering
    \begin{subfigure}[t]{0.48\linewidth}
    \centering
    \includegraphics[width=\linewidth]{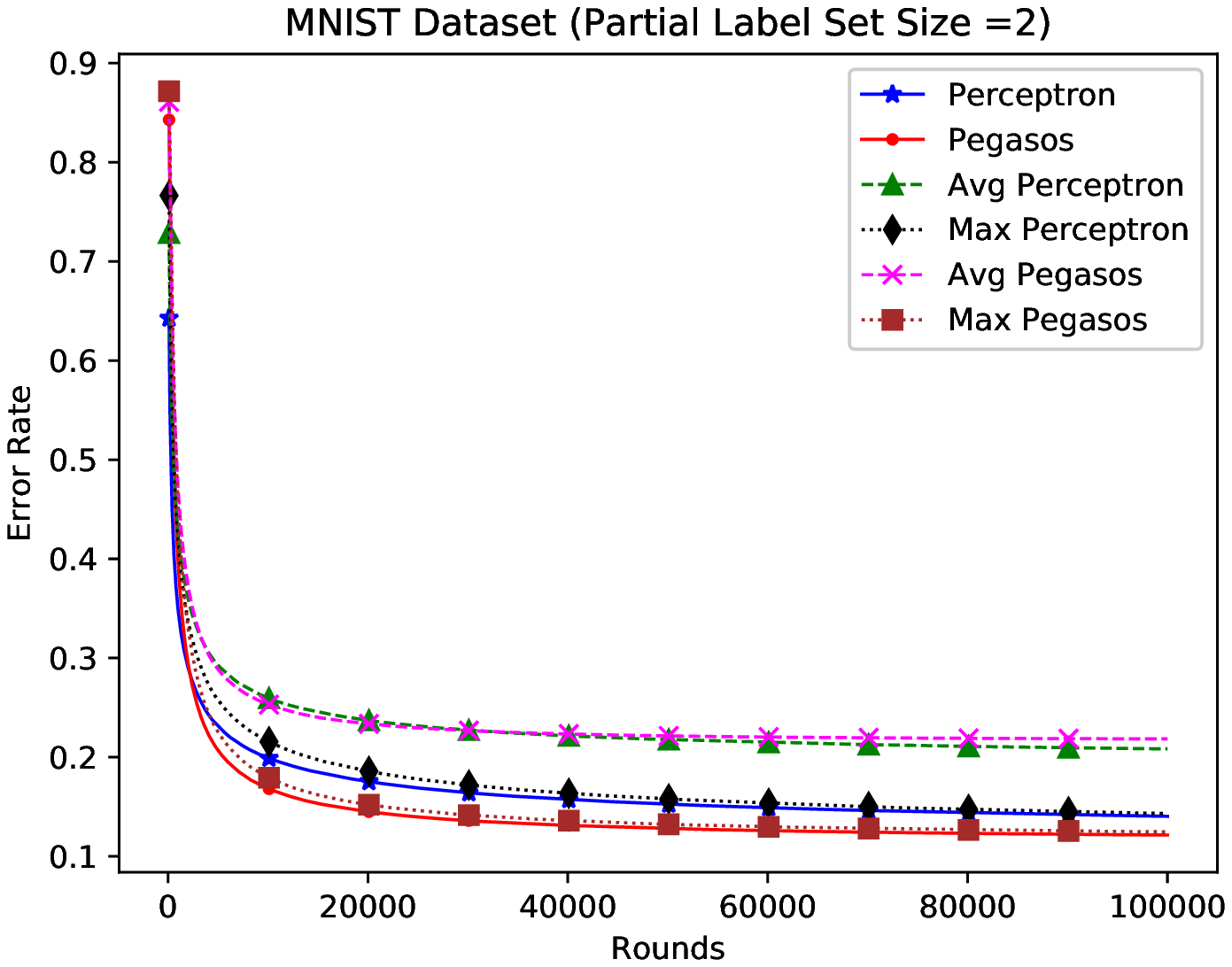}
    \end{subfigure}
    \begin{subfigure}[t]{0.48\linewidth}
    \centering
    \includegraphics[width=\linewidth]{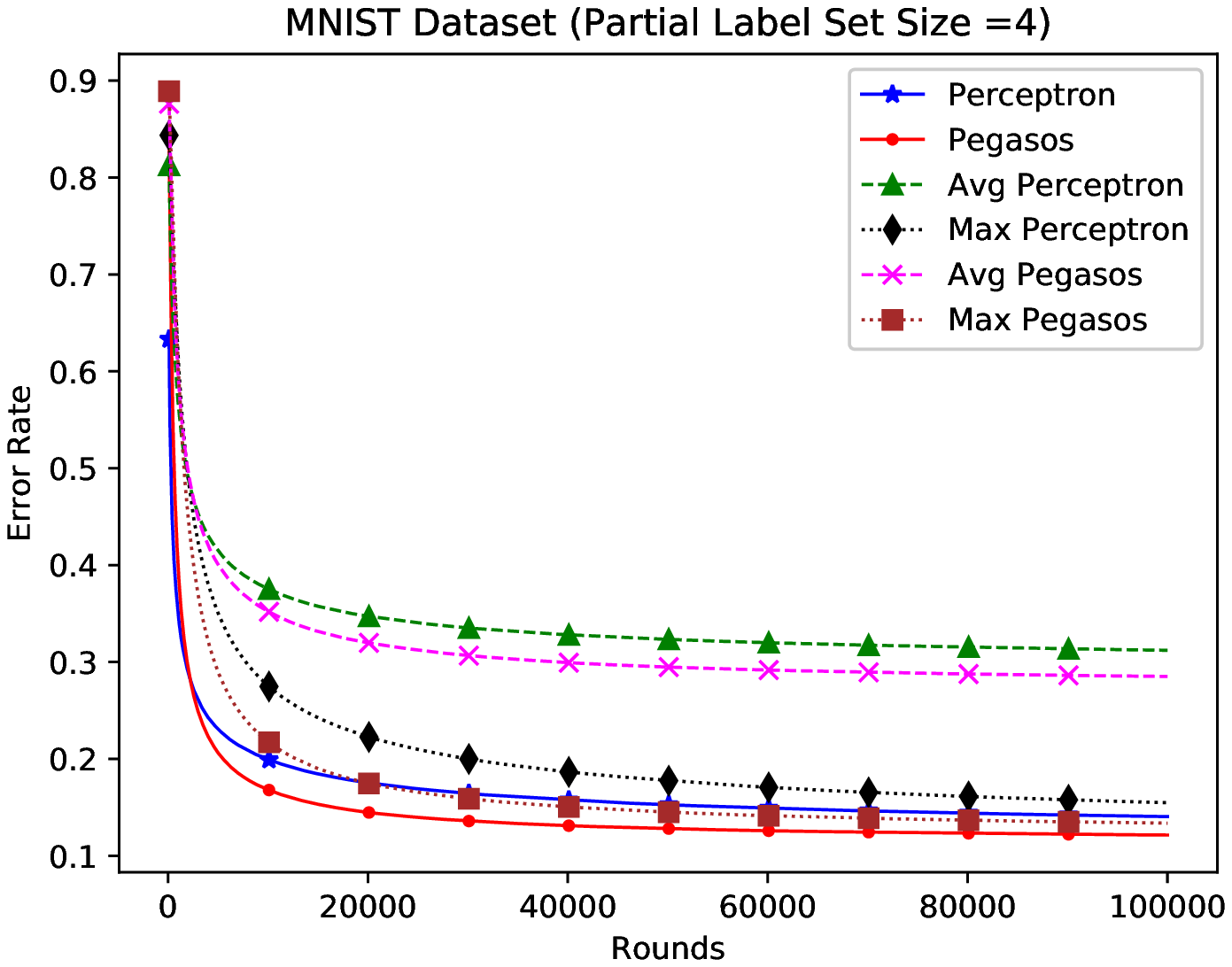}
    \end{subfigure}
    \begin{subfigure}[t]{0.48\linewidth}
    \centering
    \includegraphics[width=\linewidth]{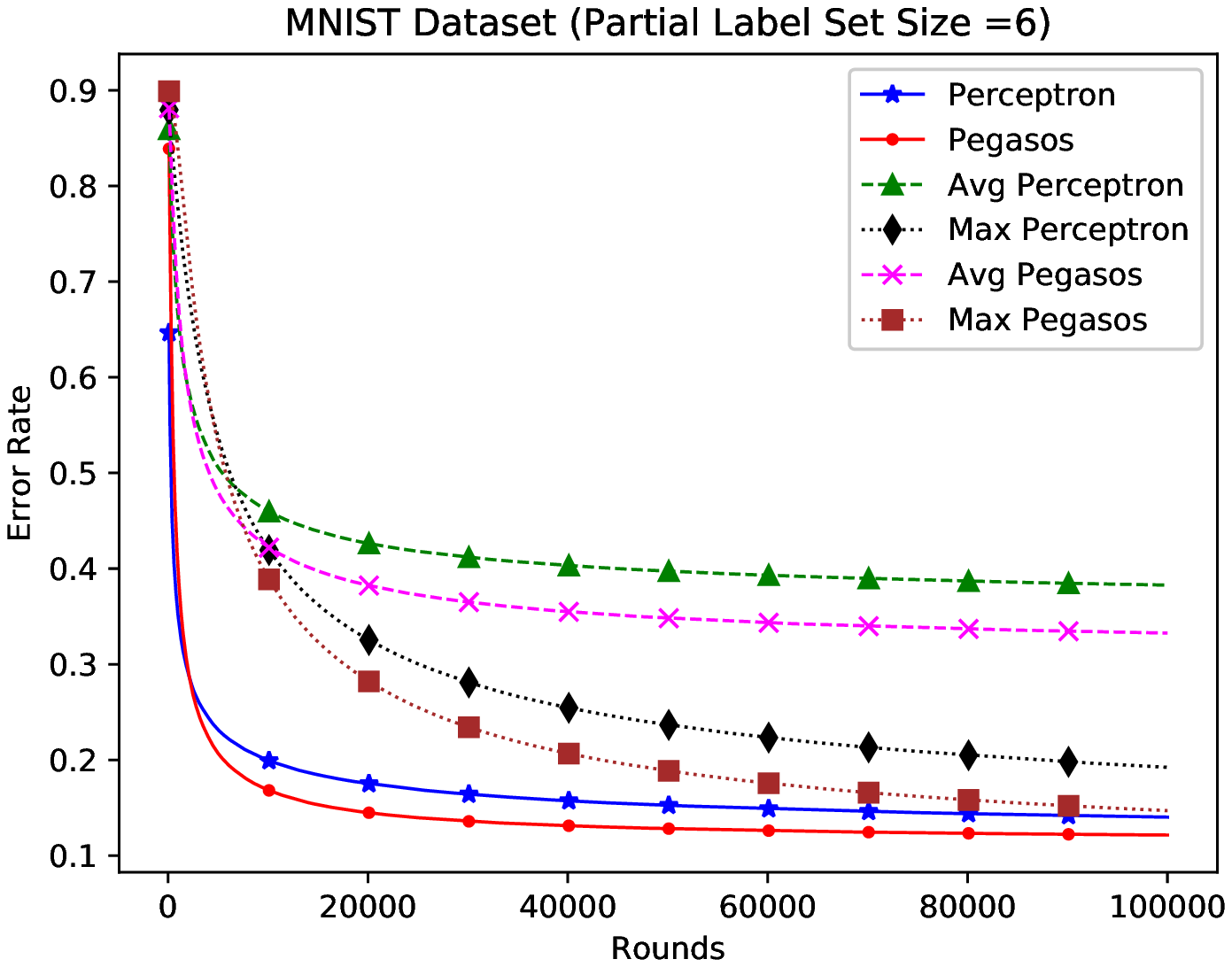}
    \end{subfigure}
    \begin{subfigure}[t]{0.48\linewidth}
    \centering
    \includegraphics[width=\linewidth]{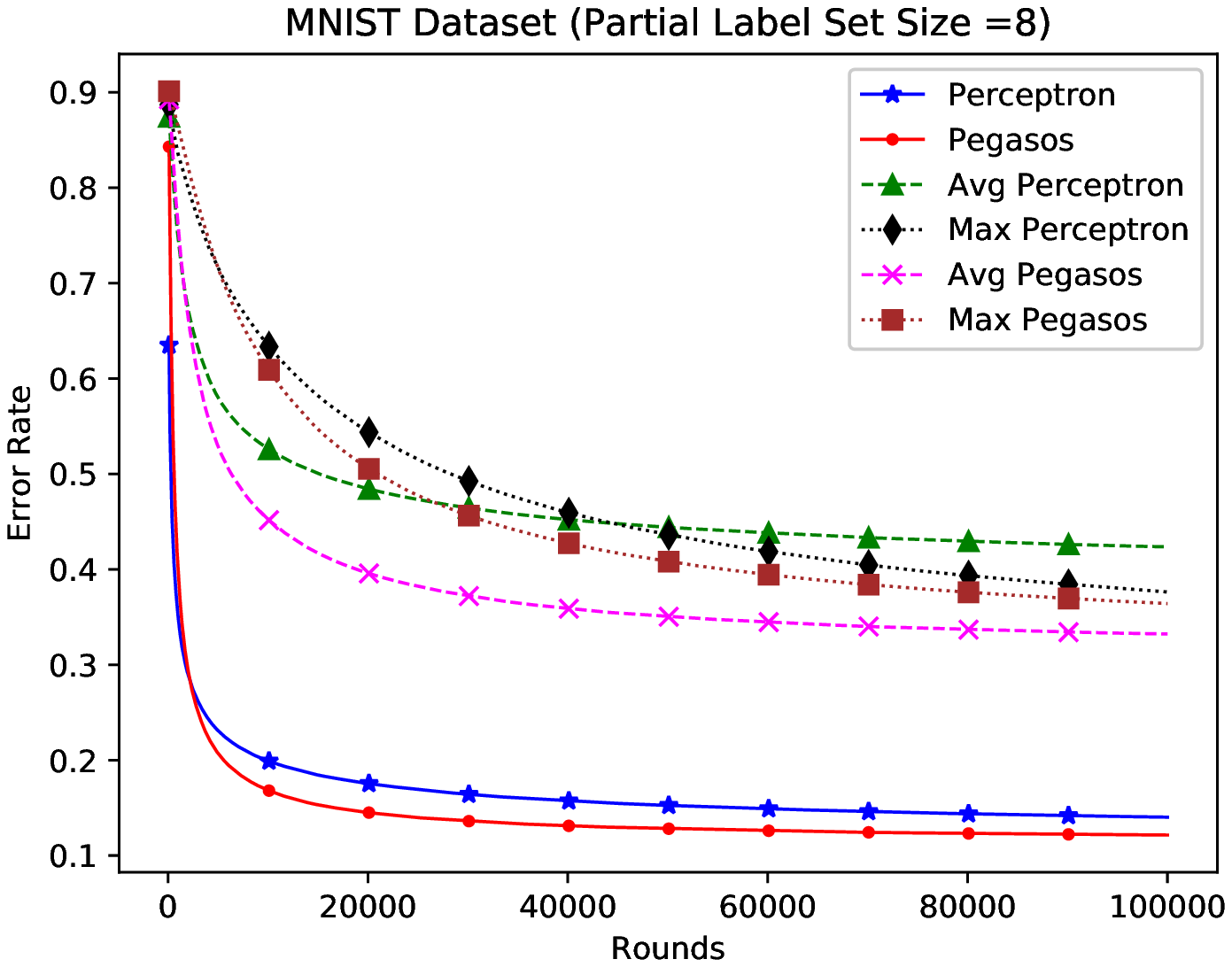}
    \end{subfigure}
    \caption{MNIST Dataset Results}
    \label{fig:mnist}
\end{figure*}
The results for Satimage and USPS datasets are shown in Fig. \ref{fig:satimage} and \ref{fig:usps} respectively. For Satimage, the Max Pegasos performs the best for label set of size 2. But for label set size 4, the Average Prediction Loss based algorithms perform much better. For USPS, we see that though for candidate labels set sizes 2 and 4, the Max Perceptron and Max Pegasos perform better than our algorithms, for label set sizes 6 and 8, the Average Prediction Loss based algorithms perform much better. The results for MNIST are provided in Fig. \ref{fig:mnist}. Here we observe the Max Perceptron and Max Pegasos performs much better than the other algorithms for label set sizes 2 and 4. However, for label set sizes 6 and 8, the Average Pegasos performs best.

Overall, we see that for smaller labels set sizes, the Max Prediction Loss performs quite well. However, the Average Prediction Loss shows the best for larger candidate label set sizes. Studying the convergence and theoretical properties of the non-convex Max Prediction Loss can be an exciting future direction for exploration.

\section{Conclusion}
In this paper, we proposed online algorithms for classifying partially labeled data. This is very useful in real-life scenarios when multiple annotators give different labels for the same instance. We presented algorithms based on the Perceptron and Pegasos. We also provide mistake bounds for the Perceptron based algorithm and the regret bound for the Pegasos based algorithm. We also provide an experimental comparison of all the algorithms on various datasets. The results show that though the Average Prediction Loss is convex, the non-convex Max Prediction Loss can also be useful for small labels set sizes. 
Providing a theoretical analysis for the Max Prediction Loss can be a useful endeavor in the future.

\newpage
\begin{appendix}

\section{Proof of Theorem~1}
\label{proof-thm1}
\begin{proof}
Assume that at the round $t$, the algorithm fails to classify $(\mathbf{x}^t,Y^t)$ with the proper margin using the weight matrix $W^t$, that is, $\frac{1}{|Y^t|}\sum_{i\in Y^t}\mathbf{w}_i^t.\mathbf{x}^t -\max_{j\in \overline{Y}^t}\mathbf{w}_j^t.\mathbf{x}^t <1$ or $L(h^t(\mathbf{x}^t),Y^t)>0$. So, the weights are updated using the rule $\mathbf{w}_i^{t+1}=\mathbf{w}_i^t+\tau_i^t\mathbf{x}^t,\; i \in [K]$ where the $\tau_i^t$ are as specified in Algorithm 1. To prove the theorem, we bound $||W^T||_2^2$ from above and below. First, we derive the lower bound for $\sum_{i=1}^K\mathbf{w}^*_i.\mathbf{w}^{t+1}_i$.
\begin{align}
    \nonumber & \sum_{i=1}^K\mathbf{w}^*_i.\mathbf{w}^{t+1}_i = \sum_{i=1}^K\mathbf{w}^*_i.(\mathbf{w}^t_i+\tau_i^t\mathbf{x}^t) \\ 
   \nonumber  &= \sum_{i=1}^K\mathbf{w}^*_i.\mathbf{w}^{t}_i + \sum_{i=1}^K\tau_i^t(\mathbf{w}^*_i.\mathbf{x}^t) \\
     \nonumber &= \sum_{i=1}^K\mathbf{w}^*_i.\mathbf{w}^{t}_i+\frac{1}{|Y^t|}\sum_{i \in Y^t}\mathbf{w}^{*}_i.\mathbf{x}^t-\max_{j \in  \overline{Y}^t}\mathbf{w}^{*}_j.\mathbf{x}^t\\
     &\geq \sum_{i=1}^K\mathbf{w}^*_i.\mathbf{w}^{t}_i+\gamma \mathbb{I}_{\{\frac{1}{|Y^t|}\sum_{i\in Y^t}\mathbf{w}_i^t.\mathbf{x}^t -\max_{j\in \overline{Y}^t}\mathbf{w}_j^t.\mathbf{x}^t <1\}} \label{eq:1}
\end{align}
We get the above expression due to the assumption that $W^*$ classifies all points with margin at least $\gamma$. Summing Eq.(\ref{eq:1}) from $t=1$ to $T$, we get the following. Thus, if the algorithm made $m$ mistakes in $T$ trials, we get.
\begin{align}
  \nonumber & \sum_{t=1}^T \sum_{i=1}^K\mathbf{w}^*_i.\mathbf{w}^{t+1}_i \geq \sum_{t=1}^T\sum_{i=1}^K\mathbf{w}^*_i.\mathbf{w}^{t}_i\\
  \nonumber &+ \gamma\sum_{t=1}^T \mathbb{I}_{\{\frac{1}{|Y^t|}\sum_{i\in Y^t}\mathbf{w}_i^t.\mathbf{x}^t -\max_{j\in \overline{Y}^t}\mathbf{w}_j^t.\mathbf{x}^t <1\}}\\
  \nonumber \Rightarrow &  \sum_{i=1}^K\mathbf{w}^*_i.\mathbf{w}^{T+1}_i \geq \sum_{i=1}^K\mathbf{w}^*_i.\mathbf{w}^{1}_i+ \gamma m\geq \gamma m\\
  \Rightarrow &  W^*.W^{T+1} \geq \gamma m\label{eq:lower-bound}
\end{align}
Where we used the fact that  and $W^1=\mathbf{0}_{d\times K}$. Let $W^*.W^{T+1}$ be the Frobenius inner product between $W^*$ and $W^{T+1}$. Then, using Cauchy-Schwartz inequality, we get the following. 
\begin{align}
\nonumber     (W^*.W^{T+1})^2&=(\sum_{i=1}^K\mathbf{w}^*_i.\mathbf{w}^{T+1}_i)^2 \leq \sum_{i=1}^K \Vert \mathbf{w}^*_i\Vert_2^2 .\Vert \mathbf{w}^{T+1}_i\Vert_2^2 \\
    \nonumber &\leq (\sum_{i=1}^K \Vert \mathbf{w}^*_i\Vert_2^2) (\sum_{i=1}^K \Vert \mathbf{w}^{T+1}_i\Vert_2^2)\\
    &=\Vert W^* \Vert^2. \Vert W^{T+1}\Vert^2 \label{eq:upper-bound}
\end{align}
From Eq.(\ref{eq:lower-bound}) and (\ref{eq:upper-bound}) and using the assumption that  $||W^*||=1$, we get:
\begin{equation}
\label{eq:lower_bound-1}
    ||W^{T+1}||^2 \geq m^2\gamma^2
\end{equation}
Now, we derive upper bound on $||W^T||$. We know that at $t^{th}$ trial, example $\mathbf{x}^t$ is misclassified. Thus,
\begin{equation}\label{upper_start}
\begin{split}
    ||W^{t+1}||^2 &= \sum_{i=1}^K||\mathbf{w}_i^{t+1}||^2 =\sum_{i=1}^K||\mathbf{w}_i^t+\tau_i^t\mathbf{x}^t||^2 \\
    &= \sum_{i=1}^K||\mathbf{w}_i^t||^2+2\sum_{i=1}^K\tau_i^t(\mathbf{w}_i^t.\mathbf{x}^t)+\sum_{i=1}^K||\tau_i^t\mathbf{x}^t||^2 \\
    &= \sum_{i=1}^K||\mathbf{w}_i^t||^2+2\sum_{i=1}^K\tau_i^t(\mathbf{w}_i^t.\mathbf{x}^t)+||\mathbf{x}^t||^2\sum_{i=1}^K(\tau_i^t)^2.
\end{split}
\end{equation}
Using $||\mathbf{x}^t|| \leq R$, $
    \sum_{i=1}^K\tau_i^t \mathbf{w}_i^t.\mathbf{x}^t = \frac{1}{|Y^t|}\sum_{i \in Y^t}\mathbf{w}^{t}_i.\mathbf{x}^t-\max_{j \in  \overline{Y}^t}\mathbf{w}^{t}_j.\mathbf{x}^t < 1$
and $\sum_{i=1}^L(\tau_i^t)^2 =\frac{1}{|Y^t|}+1$ in  Eq.(\ref{upper_start}), we get the following.
\begin{equation*}
\begin{split}
    &||W^{t+1}||^2-||W^t||^2 \\
    &\leq \left(2+\left[\frac{1}{|Y^t|}+1\right]R^2\right)\mathbb{I}_{\{\frac{1}{|Y^t|}\sum_{i\in Y^t}\mathbf{w}_i^t.\mathbf{x}^t -\max_{j\in \overline{Y}^t}\mathbf{w}_j^t.\mathbf{x}^t <1\}}
\end{split}
\end{equation*}
We know that $\Vert W^1\Vert^2=0$ and there are $m$ mistakes. Summing the above equation over $t=1$ to $T$, we get,
\begin{align}
   \nonumber  &\Vert W^{T+1}\Vert^2- \Vert W^1 \Vert^2 \leq 2m+\left[\frac{1}{c}+1\right]mR^2\\
    \Rightarrow & \Vert W^{T+1}\Vert^2 \leq 2m+\left[\frac{1}{c}+1\right]mR^2. \label{upper_bound}
\end{align}
Where, $c=\min_{t}|Y^t|$. Thus, combining the upper and lower bound from Eq.(\ref{eq:lower_bound-1}) and (\ref{upper_bound}), we get the following.
\begin{align*}
    & m^2\gamma^2 \leq ||W^{T+1}||^2 \leq 2m+\left[\frac{1}{c}+1\right]mR^2 \nonumber\\
    \Rightarrow & m \leq \frac{2}{\gamma^2}+\left[\frac{1}{c}+1\right]\frac{R^2}{\gamma^2} \nonumber
\end{align*}
\end{proof}

\section{Proof of Theorem~2}
\label{proof-thm2}
\begin{proof}
If $D=0$, it reduces to linearly separable case and thus, we assume $D>0$. Which means, there exists $t\in [T]$ such that $ d^t=\max \{0,\gamma-[\frac{1}{|Y^t|}\sum_{i \in Y^t}\mathbf{w}_i.\mathbf{x}^t-\max_{j \in  \overline{Y}^t}\mathbf{w}_j.\mathbf{x}^t]\}>0$. Thus, the data is not linearly separable with respect to $W$. We now transform the linearly non-separable data to separable data. We extend each instance $\mathbf{x}^t \in \mathbb{R}^d$ to $\mathbf{z}^t \in \mathbb{R}^{d+T}$ as follows. The first d coordinates of $\mathbf{z}^t$ are set to $\mathbf{x}^t$. The $(d+t)$th coordinate of $\mathbf{z}^t$ is set to $\Updelta$ whose value will be determined later while the rest of the coordinates of $\mathbf{z}^t$ are set to 0. We extend weight matrix $W$ to $M \in \mathbb{R}^{(d+T)\times K}$ as follows. We set the first d columns of $M$ to be $\frac{1}{Z}W$ (where Z is a constant whose value will be determined). For the rest of the columns, we set the $(d+t,t)^{th}$ position in $M$ to $\frac{d^t}{Z}\Updelta$ if $r \in Y^t$ and to 0 otherwise.

We choose the value of $Z$ such that $||M||_2=1$ and hence,
\begin{equation}
    1=||M||_2^2=\frac{1}{Z^2}\left(||W||_2^2+\frac{D^2}{\Updelta^2}\right). \nonumber
\end{equation}
This gives us,
\begin{equation}
    Z=\sqrt{1+\frac{D^2}{\Updelta^2}}. \nonumber
\end{equation}
Let $\mathbf{m}_r$ be the $r^{th}$ column of $M$, then 
$\mathbf{m}_r.\mathbf{x}^t = \frac{1}{Z}\left(\mathbf{w}_r.\mathbf{x}^t+\mathds{I}_{r \in Y^t}\frac{d^t}{\Updelta}\Updelta\right)$. We now show that $M$ linearly separates all the examples $\mathbf{z}^t$ with a margin at least $\frac{\gamma}{Z}$ as follows. 
\begin{equation*}
\begin{split}
    &\frac{1}{|Y^t|}\sum_{i \in Y^t}\mathbf{m}_i.\mathbf{x}_t-\max_{j \in  \overline{Y}^t}\mathbf{m}_j.\mathbf{x}_t \\
    &= \frac{1}{Z|Y^t|}\sum_{i \in Y^t}(\mathbf{w}_i.\mathbf{x}_t+d^t)-\max_{j \in  \overline{Y}^t}\{\frac{1}{Z}\mathbf{w}_j.\mathbf{x}_t\} \\
    &= \frac{1}{Z}d^t+\frac{1}{Z}\left[\frac{1}{|Y^t|}\sum_{i \in Y^t}\mathbf{w}_i.\mathbf{x}_t-\max_{j \in  \overline{Y}^t}\mathbf{w}_j.\mathbf{x}_t\right] \\
    &\geq \frac{1}{Z}(\gamma-[\frac{1}{|Y^t|}\sum_{i \in Y^t}\mathbf{w}_i.\mathbf{x}_t-\max_{j \in  \overline{Y}^t}\mathbf{w}_j.\mathbf{x}_t]) \\
    &+\frac{1}{Z}\left[\frac{1}{|Y^t|}\sum_{i \in Y^t}\mathbf{w}_i.\mathbf{x}_t-\max_{j \in  \overline{Y}^t}\mathbf{w}_j.\mathbf{x}_t\right] = \frac{\gamma}{Z}
\end{split}
\end{equation*}
We also observe that
$ ||\mathbf{z}^t||_2^2 = ||\mathbf{x}^t||_2^2+\Updelta^2 \leq R^2+\Updelta^2$.
Thus, using Theorem~\ref{linear_mistake}, the number of mistakes made by the algorithm Avg Perceptron on the sequence $(\mathbf{z}^1,Y^1),\ldots,(\mathbf{z}^T,Y^T)$ is bounded above as follows.
\begin{equation}\label{nonlinear_bound_1}
    m \leq 2\frac{Z^2}{\gamma^2}+2\left[\frac{1}{c}+1\right]\frac{R^2+\Updelta^2}{(\frac{\gamma}{Z})^2}
\end{equation}
Minimizing RHS expression in Eq.(\ref{nonlinear_bound_1}) over $\Updelta$, we get that the optimal value of $\Updelta$ is $\left[\frac{D^2+KD^2R^2}{K}\right]^{\frac{1}{4}}$ where $K=1+\frac{1}{c}$. Using this value of $\Updelta$, we get the mistake bound as follows.
\begin{equation}
     m\leq 2\frac{Z^2}{\gamma^2}+2K\frac{R^2+\Updelta^2}{(\frac{\gamma}{Z})^2} \nonumber
\end{equation}

Finally, to complete the proof we need to show that classifying the original partially labeled sequence with matrices $W^1,\ldots,W^T$ is the same as classifying as the extended sequence with the extended matrices $M^1,\ldots,M^T$. That is, they both produce same sequence of predictions. This can be accomplished if we can show the following holds for all $t\in [T]$.
\begin{enumerate}
\item The first d columns of $M^t$ are equal to $W^t$ 
\item The (d+t)th column of $M^t$ is zero.
\item $ \mathbf{m}_r^t.\mathbf{x}^t=\mathbf{w}_r^t.\mathbf{x}^t \quad \forall r \in \{1,2,...,L\}$
\end{enumerate}
The proof of the above conditions is straightforward by induction on t (by initializing $M^1$ and $W^1$ as zero matrices).
\end{proof}

\section{Proof of Theorem~3}
\label{proof-thm3}
\begin{proof}
The theorem and the proof is almost same as Theorem 1 and its proof in the Pegasos paper \cite{Pegasos}. The main idea in the proof is to upper bound $||\nabla^t||$ where $\nabla^t$ is given by Eq. \ref{pegasos_grad}. Thus, using triangle inequality we can write:
\begin{equation}\label{bound}
    ||\nabla^t||\leq \lambda||W^t||+||\nabla_{W^t}L||
\end{equation}
We note that the L2 norm of the weight matrix $W^t$ can be written as $||W^t||^2=\sum\limits_{i=1}^k||\mathbf{w}^t_i||^2$. Now, $||W^t||\leq \frac{1}{\sqrt{\lambda}}$ and $||\nabla_{W^t}L||^2=\sum\limits_{i=1}^k||\nabla_{\mathbf{w}^t_i}L||^2$. From the updates of Avg Perceptron, we get:
\begin{equation}
    ||\nabla_{W^t}L||^2=\begin{cases}
    ||\mathbf{x}^t||^2+\frac{||\mathbf{x}^t||^2}{|Y^t|}, & \text{if } L>0 \\
    0, & \text{if } L=0 \nonumber
    \end{cases}
\end{equation}
So we get, 
\begin{equation}
||\nabla_{W^t}L|| \leq \sqrt{1+\frac{1}{Y^t}}||\mathbf{x}^t|| \nonumber
\end{equation}
So, using the above result along with Equation \ref{bound}, we can write:
\begin{equation}
    ||\nabla^t|| \leq \sqrt{\lambda}+\sqrt{1+\frac{1}{Y^t}}||\mathbf{x}^t|| \nonumber
\end{equation} 
Thus, if $c=\min_t |Y^t|$, we get the following bound:
\begin{equation}
    ||\nabla^t|| \leq \sqrt{\lambda}+\sqrt{1+\frac{1}{c}}||\mathbf{x}^t|| \nonumber
\end{equation} 
The rest of the proof is exactly same as the one given in \cite{Pegasos}. 
\end{proof}

\end{appendix}
%
%
%
%
\bibliography{PL}
\bibliographystyle{splncs04}

\end{document}